\documentclass[runningheads]{llncs}
\usepackage[colorlinks=true, allcolors=black, citecolor=blue]{hyperref} 
\usepackage{tablefootnote} 
\usepackage{tabularx}
\usepackage{subfig}
\usepackage{graphicx}
\begin{document}
\title{FairytaleQA Translated: Enabling Educational Question and Answer Generation in Less-Resourced Languages\thanks{Preprint - Accepted for publication at ECTEL 2024.}
}

\titlerunning{}
%
\author{
Bernardo~Leite\inst{}\orcidID{0000-0002-9054-9501} \and 
Tom\'as~Freitas~Os\'orio\inst{}\orcidID{0009-0001-2036-3197}
\and
Henrique~Lopes~Cardoso\inst{}\orcidID{0000-0003-1252-7515}
}

\authorrunning{B. Leite et al.}
%
\institute{
Artificial Intelligence and Computer Science Laboratory (LIACC) \\
Faculty of Engineering of the University of Porto (FEUP) \\
Rua Dr. Roberto Frias, 4200-465 Porto, Portugal \\
\email{bernardo.leite@fe.up.pt} \\
\email{tomas.s.osorio@gmail.com} \\
\email{hlc@fe.up.pt}
}
\maketitle              
\begin{abstract}

Question Answering (QA) datasets are crucial in assessing reading comprehension skills for both machines and humans.
While numerous datasets have been developed in English for this purpose, a noticeable void exists in less-resourced languages. 
To alleviate this gap, our paper introduces machine-translated versions of FairytaleQA, a renowned QA dataset designed to assess and enhance narrative comprehension skills in young children. By employing fine-tuned, modest-scale models, we establish benchmarks for both Question Generation (QG) and QA tasks within the translated datasets.
In addition, we present a case study proposing a model for generating question-answer pairs, with an evaluation incorporating quality metrics such as question well-formedness, answerability, relevance, and children suitability.
Our evaluation prioritizes quantifying and describing error cases, along with providing directions for future work. This paper contributes to the advancement of QA and QG research in less-resourced languages, promoting accessibility and inclusivity in the development of these models for reading comprehension. The code and data is publicly available at \url{github.com/bernardoleite/fairytaleqa-translated}.

\end{abstract}
\section{Introduction}

Question Answering (QA)\footnote{Henceforth, we use ``QA dataset'' to denote datasets comprising questions and answers, and ``QA task'' to refer to the computational task of answering a question.} datasets, typically composed of text passages and corresponding question-answer pairs, have been used in advancing Question Generation (QG) and QA tasks research. 
Building these datasets often requires extensive effort from domain experts who are well-versed in the specific domain being targeted. A notable example is the FaitytaleQA \cite{xu_2022_fairytaleqa}, a high-quality QA dataset carefully designed by education experts who applied evidence-based narrative comprehension frameworks \cite{paris_2003_narrative,alonzo_2009_narrative} to annotate each question. While FairytaleQA excels in its quality and design, its content is exclusively tailored to a single language --- English. This is also the case in other QA datasets, which presents a challenge for researchers aiming to extend their research to other languages. Notably, this challenge applies to languages that, despite not being low-resourced, lack QA datasets of this nature (referred to here as \textit{less}-resourced languages).
One viable solution involves relying on machine translation to generate reliable datasets, namely those that support the evaluation of neural models in downstream tasks. While machine translation is still not without issues, the resulting datasets can serve the purpose of fostering research in less-resourced languages.

\begin{figure}[htp]
    \centering
    \includegraphics[scale=0.46]{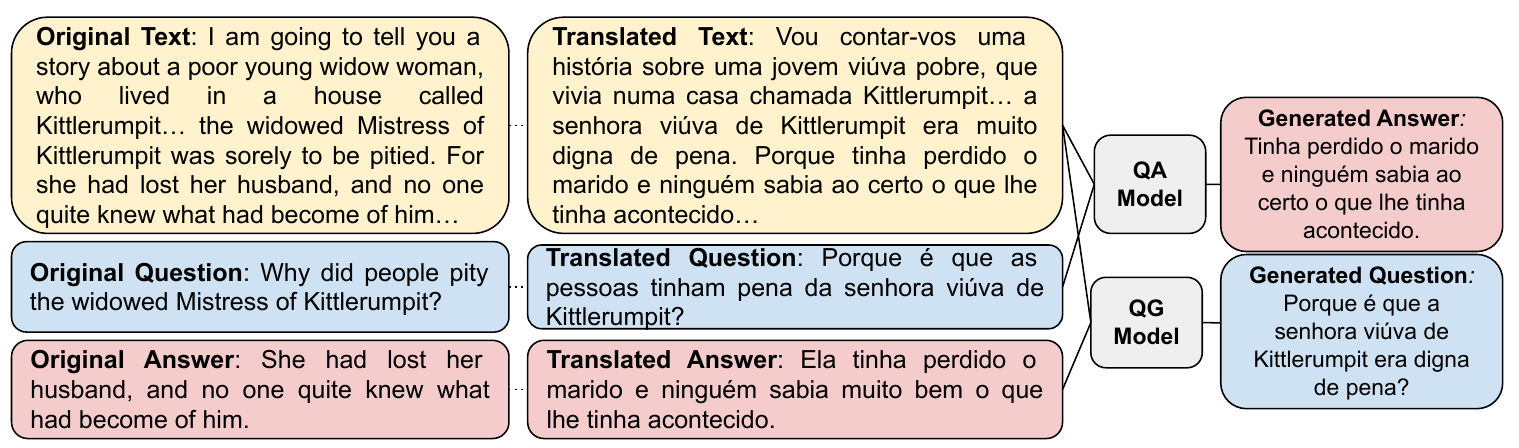}
    \caption{QA and QG tasks using a translated text and corresponding QA pair.} 
    \label{fig:overview_qg_qa}
\end{figure}

Motivated by this observation, the first goal of this work is to provide machine-translated versions of FairytaleQA, in three\footnote{We also have included translated datasets for Italian and Romanian in our repository, although they were not studied in this research.} of the most widely spoken Romance languages: Spanish, Portuguese (European pt-PT and Brazilian pt-BR), and French. To estimate the quality of the translated versions, we conduct an error analysis of one of the translated datasets.
As our second goal, we establish baseline benchmarks for both QA and QG tasks, achieved through fine-tuning modest-scale models. These models offer the advantage of being usable on more affordable hardware, in contrast to large language models that might incur costs for using expensive hardware or external APIs. Figure~\ref{fig:overview_qg_qa} shows an example where the original text and a QA pair are translated into Portuguese and applied in both QA and QG tasks.

The third goal entails our case study, which focuses on presenting and evaluating a model designed for Question-Answer Pair Generation (QAPG), specifically targeting the Portuguese language (pt-PT).
Our motivation for generating both questions and answers is driven by the need for automated assessment of student answers in real-world educational scenarios.
We aim particularly to understand the faithfulness of QA pairs generated by a modest-scale QAPG, trained on translated data, and compare them to those found in actual exams. This forms the central research question of our case study:

RQ: \textit{Can a modest-scale QAPG model, trained on translated data, generate QA pairs that are qualitatively similar to those used in real exams in a less-resourced language?}

We conduct human evaluation based on criteria such as well-formedness, answerability, relevance, and suitability for children. While our case study focuses on a single language, we expect it can serve as (1) a guideline for researchers seeking to explore QA/QG in other translated datasets, and (2) a representative illustration of expected errors during QA pair generation, along with directions for future work.

Our analysis found that the QAPG model trained on translated data can generate well-formed questions. However, detection of semantic ambiguities was noted. Additionally, our observations revealed that the alignment of the generated QA pairs is not always consistent, indicating the need for further refinement.
In summary, our contributions are:
\begin{itemize}
    \item We provide machine-translated versions\footnote{Made available at \url{github.com/bernardoleite/fairytaleqa-translated}.} of the FairytaleQA dataset in three widely spoken Romance languages.
    \item We introduce fine-tuned models\footnote{Made available at \url{github.com/bernardoleite/fairytaleqa-translated}.} along with benchmarks for both QA and QG tasks within the translated datasets.
    \item We present a case study where a QAPG model is proposed and evaluated, focusing on quantifying and describing expected errors in QA pair generation.
\end{itemize}

\section{Background and Related Work}

\subsection{Question \& Answer Generation Corpora}

Over the years, numerous QA and QG datasets\footnote{QA datasets are typically also used for QG.} have been proposed. A reference example is the SQuAD dataset \cite{rajpurkar_2016_squad}, comprising crowd-sourced QA pairs extracted from Wikipedia articles.
In the case of SQuAD, to assess its viability in other languages, some machine-translated versions have been made publicly available, including Spanish \cite{carrino_2020_qa_spanish} and Slovak \cite{stavs_2023_squadslovak}.
However, a limitation of these datasets is that they are built upon open-domain and general-purpose resources like Wikipedia and news articles. As a result, models trained on this type of data (or translations) may generate suitable questions/answers for a generic domain but are likely to fall short in serving specific educational purposes.

Focusing on corpora explicitly created for an educational purpose, RACE \cite{lai_2017_race}, CLOTH \cite{xie_2018_cloth}, and FairytaleQA \cite{xu_2022_fairytaleqa} are prime examples of this accomplishment. The questions within RACE and CLOTH were derived from English exams aimed at middle to high school students. Notably, CLOTH labels each question with its corresponding reasoning level, providing a valuable advantage for educational applications.

\subsubsection{Why FairytaleQA?}
We selected the FairytaleQA dataset due to its texts and corresponding QA pairs aligning with the goal of \textit{supporting narrative comprehension}. As highlighted by Xu \textit{et al.} \cite{xu_2022_fairytaleqa}, narrative comprehension is a high-level skill that strongly correlates with reading success \cite{lynch_2008_reading}. Furthermore, narrative stories possess a well-defined structure comprising distinct elements and their relationships.
We also selected this dataset due to its structured focus on specific dimensions of reading comprehension skills. Specifically, education experts have annotated each question, following evidence-based narrative comprehension frameworks \cite{paris_2003_narrative,alonzo_2009_narrative}, and addressing two key attributes: \textit{narrative elements} and \textit{explicitness} (explained in Section~\ref{sec:about_dataset}). To the best of our knowledge, this is the first work involving translations of FairytaleQA.

\subsection{Question \& Answer Generation Tasks}
While the QA task revolves around finding or generating an answer for a certain question, QG involves the automatic generation of well-structured questions. Both tasks can leverage diverse data sources, including free text or knowledge bases.
QG holds practical significance in education. It can assist teachers in creating assessments or stimulate students' self-learning \cite{kurdi_2020_education}.

From a methodological standpoint, prior work on QA and QG has primarily focused on fine-tuning small to medium-sized pre-trained language models \cite{leite_2022_nqg,zhao_2022_cqg_acl,ghanem_2022_cqg_acl} for generating questions and answers. 
With the emergence of large language models \cite{brown_2020_terms}, recent research leveraged the prompting paradigm, where a textual query specifies the desired generation task given a set of prompt examples. Although this technique has demonstrated compelling performance \cite{elkins_2023_cqg_aied}, it can be reliant on hardware or external API access with associated costs, posing a limitation. In this study, we concentrate on employing modest-scale fine-tuned models. The term ``modest-scale'' is used because we specifically employ T5-based models \cite{raffel_2020_t5} with parameters in the magnitude of millions, making it suitable for more affordable hardware. This is in contrast to large language models, which typically have parameters in the billions\footnote{It is worth noting recent efforts to make these models more accessible, for instance through quantization techniques.
}. Nevertheless, we compare the performance of the case study's model (T5) against a large language model, i.e., GPT-4 (Section~\ref{sec:case_study}).

\section{Translating the FairytaleQA Dataset}

\subsection{About FairytaleQA} \label{sec:about_dataset}
FairytaleQA includes 10,580 QA pairs, manually created by educational experts based on 278 stories. 
On average, each story comprises 15 sections, and each section (paragraph), made of multiple sentences, contains an average of 3 questions. Each QA pair is annotated with corresponding narrative elements and explicitness labels. 
Narrative elements are described as follows:
\begin{itemize}
    \item \textbf{Character}: These require test takers to identify or describe the characteristics of story characters.
    \item \textbf{Setting}: These inquire about the place or time where story events occur and typically start with ``Where'' or ``When''.
    \item \textbf{Action}: These focus on characters' behaviors or seek information about their actions.
    \item \textbf{Feeling}: These explore characters' emotional status or reactions, often framed as ``How did/does/do...feel''.
    \item \textbf{Causal relationship}: These examine the cause-and-effect relationships between two events, often starting with ``Why'' or ``What made/makes''.
    \item \textbf{Outcome resolution}: These ask for the events that result from prior actions in the story, typically phrased as ``What happened/happens/has happened...after...''.
    \item \textbf{Prediction}: These request predictions about the unknown outcome of a particular event based on existing information in the text.
\end{itemize}
Question explicitness is defined as follows:
\begin{itemize}
    \item \textbf{Explicit}: The answers are directly present in the text and can be located within specific passages.
    \item \textbf{Implicit}: Answers cannot be directly pinpointed in the text, requiring the ability to summarize and make inferences based on implicit information.
\end{itemize}

\subsection{Machine Translation and Sample Evaluation} \label{sec:from_translation_to_error}

Each dataset instance contains a text section (paragraph) and a QA pair. Consequently, to translate each instance effectively, we needed to address these three components.
We opted to translate all three components together, separated by a line break. This approach was chosen based on empirical evidence from preliminary test translations, indicating that translating questions and answers with improved contextualization of the section text led to greater coherence between these components.
For machine translation, we have used DeepL\footnote{\url{https://www.deepl.com/translator}}, a commercial tool known for its translation quality compared to its competitors. We translated the original English version into the three most spoken Romance languages: Spanish, Portuguese (European pt-PT and Brazilian pt-BR), and French.

To assess the translation quality of the data, we conducted a manual error analysis on 10 translated texts and corresponding 150 QA pairs (we selected the sections with the highest number of QA pairs). Our analysis focused on the translated version in European Portuguese (pt-PT). As all translated versions are Romance languages, any identified error types are expected to be potentially applicable to these versions, serving as a cautionary note for researchers considering their utilization.
The quantification of issues (by text, question, and answer) is presented in Table~\ref{tab:error_anal_mt} and elaborated upon below.

\begin{table}
    \centering
\caption{Incidence of issues resulting from machine translation in a dataset sample.}
\label{tab:error_anal_mt}
    \begin{tabular}{|c|c|c|c|} \hline 
 \textbf{Issue}& \textbf{Nr. Texts}& \textbf{Nr. Questions}&\textbf{Nr. Answers}\\ \hline 
         Translating Names&  7/10&  22/150& 10/150\\ \hline  
         Change of Gender&  0/10&  1/150& 2/150\\ \hline  
 Lost in Translation& 0/10& 1/150&0/150\\ \hline 
 Outdated Spelling Agreement& 1/10& 1/150&0/150\\ \hline
    \end{tabular}
\end{table}

\subsubsection{Translating Names:} A common issue arises from the non-deterministic translation of proper or common names. Take, for instance, the example ``\textit{Tom} had a little more trouble with him.'' which translates to ``O \textit{Tomás} teve um pouco mais de problemas com ele.'' This undesired translation of the character's name can have implications, especially when the name is translated in the question/answer but not in the text, introducing inconsistency between components. 
A solution to this issue could involve employing an alignment approach using Named Entity Recognition models.

\subsubsection{Change of Gender:} Some cases have been detected where the translation implies a gender change. For instance, consider the QA pair: ``How will the mother feel when the horses keep getting cut into two? Angry." which translates to: ``Como se sentirá a mãe quando os cavalos continuarem a ser cortados em dois? Zangado.'' In the translation, the answer \textit{Angry} is rendered as \textit{Zangado} which is masculine, instead of \textit{Zangada}, which is the appropriate feminine form referring to \textit{mãe} (\textit{mother}).
While finding a straightforward solution to this issue is not apparent, one option could be adjusting gender terms using set linguistic rules.

\subsubsection{Lost in Translation:} We detected one case where a question's meaning changed during translation. This occurred when translating ``What was the chariot drawn by?'' which, in back-translation, resulted in ``What chariot was pulling?''. We attribute this to the translator's performance, and at the moment, a solution is not apparent for us.

\subsubsection{Outdated Spelling Agreement:} We detected one word that did not comply with the latest spelling agreement. This occurrence may vary depending on the timeliness of the data on which the translator has been created. The solution might be using regular expressions, for example, to replace the most recent terms using a local dictionary. Additionally, spell-checking could serve as a viable option.

\section{Baseline Benchmarks: Question \& Answer Generation} \label{sec:benchmarks}

In this section, we present benchmarks for both Question Answering (QA) and Question Generation (QG) tasks within FairytaleQA.

\subsection{Implementation Details} \label{sec:implementation_details}
\begin{figure}%
    \centering
    \subfloat[\centering QA setup.]{{\includegraphics[width=5.65cm]{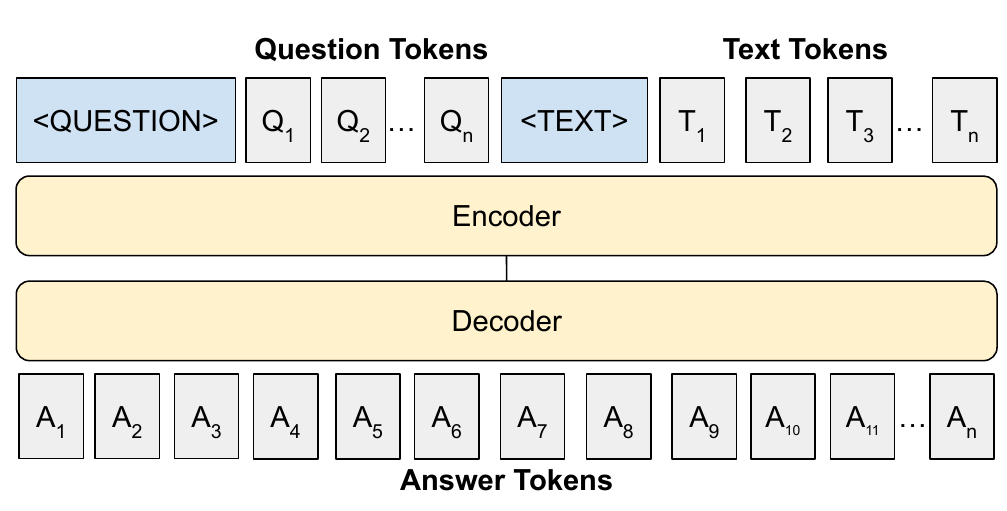} }}%
    \qquad
    \subfloat[\centering QG setup.]{{\includegraphics[width=5.65cm]{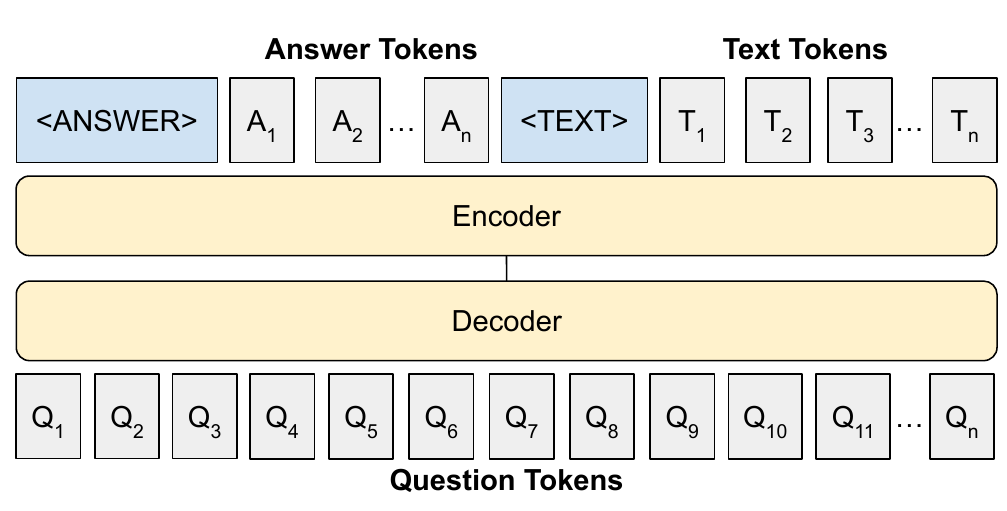} }}%
    \caption{Setups for Question Answering (QA) and Question Generation (QG).}%
    \label{fig:qaqg_setup}%
\end{figure}

Leveraging pre-trained T5 \cite{raffel_2020_t5} encoder-decoder models, known for their remarkable performance in QA/QG tasks \cite{ushio_2022_qg_t5}, we employ the \textit{t5-base}\footnote{\url{https://huggingface.co/t5-base}} version pre-trained on each language's data (monolingual). Figure~\ref{fig:qaqg_setup} shows the setup for the QA and QG tasks. For QA, the encoder concatenates the question and text, and the decoder generates the answer. In QG, the encoder concatenates the answer and text, and the decoder generates the question. We use special labels to differentiate the components.
Our maximum token input is set to 512, while the maximum token output is set to 128. During training, the models undergo a maximum of 20 epochs and incorporate early stopping with a patience of 2. A batch size of 16 is employed. During inference, we utilize beam search with a beam width of 5. 
We use the original train/val/test splits that contain 8,548/1,025/1,007 QA pairs.

\subsection{Results}
Table~\ref{tab:qa_qg_benchmarks} displays the results obtained on the test set, reported using the ROUGE$_L$-F1 metric employed by the FairytaleQA authors. It computes the \textit{n}-gram similarity (0-1) between the ground-truth questions/answers and the generated questions/answers.
\begin{table}[!ht]
\centering
\caption{QA and QG benchmarks according to the language of the translated dataset. Values are obtained through ROUGE$_L$-F1 metric (0-1).}
\label{tab:qa_qg_benchmarks}
\begin{tabular}{cc|c|ccc|}
\cline{3-6}
                                         & \multicolumn{1}{l|}{} & \textbf{QG}      & \multicolumn{3}{c|}{\textbf{QA}}                                                                   \\ \hline
\multicolumn{1}{|c|}{\textbf{Language}}  & \textbf{Model}        & \textbf{Overall} & \multicolumn{1}{c|}{\textbf{Overall}} & \multicolumn{1}{c|}{\textbf{Explicit}} & \textbf{Implicit} \\ \hline
\multicolumn{1}{|c|}{English}            & BART\tablefootnote{This model and consequent results are those reported by the authors of FairytaleQA.} \cite{lewis_2020_bart}    & 0.527            & \multicolumn{1}{c|}{0.536}            & \multicolumn{1}{c|}{0.620}             & 0.286             \\ \hline
\multicolumn{1}{|c|}{English}            & T5 \cite{raffel_2020_t5}                    & 0.530            & \multicolumn{1}{c|}{0.551}            & \multicolumn{1}{c|}{0.679}             & 0.167             \\ \hline
\multicolumn{1}{|c|}{Spanish}            & T5S \cite{araujo_2023_t5s}                   & 0.445            & \multicolumn{1}{c|}{0.382}            & \multicolumn{1}{c|}{0.474}             & 0.111             \\ \hline
\multicolumn{1}{|c|}{Portuguese (pt-PT)} & PTT5 \cite{carmo_2020_ptt5}                  & 0.496            & \multicolumn{1}{c|}{0.436}            & \multicolumn{1}{c|}{0.520}             & 0.185             \\ \hline
\multicolumn{1}{|c|}{Portuguese (pt-BR)} & PTT5 \cite{carmo_2020_ptt5}                 & 0.470            & \multicolumn{1}{c|}{0.448}            & \multicolumn{1}{c|}{0.534}             & 0.192             \\ \hline
\multicolumn{1}{|c|}{French}             & T5-fr \cite{github_2020_t5f}                   & 0.404            & \multicolumn{1}{c|}{0.431}            & \multicolumn{1}{c|}{0.518}             & 0.172             \\ \hline
\end{tabular}
\end{table}
Our overall results for QG in the original English language slightly exceed those reported by the authors of FairytaleQA (0.527 vs. 0.530), as do for overall QA (0.536 vs. 0.551). This indicates that our experimental setup aligns with the expected outcomes.
It should be noted that the values obtained in QA (Implicit) consistently appear lower than those for QA (Explicit). This discrepancy is consistent with the observations made by FairytaleQA's authors, as questions requiring implicit answers necessitate inference or summarization from the source text.

For QG, all the translated versions produced lower values than the English baseline, with statistically significant differences (t-test with \textit{$p$} $<$ .05). There is also statistical significance when comparing only among the translated versions, although the effect size is not large (Cohen’s \textit{d} $<$ 0.44).
For the QA task, we observe similar trends. However, the results of QA reveal promising evidence when delving into the explicitness aspect. As seen in the original English version, there is a substantial difference between explicit and implicit QA pairs, indicating that the translated QA pairs maintain this consistency in explicitness.

To assess the impact of translation on model performance, we decided to back-translate the Spanish dataset, which exhibited the poorest QA performance, into English. The newly translated version in English achieved the following ROUGE$_L$-F1 results: 0.497 (QG) and 0.478 (QA, overall). Notably, the QA result surpassed the performance of previous translated versions and approximated the level of the original English version\footnote{While the difference is statistically significant, the effect size is small (Cohen's \textit{d} $<$ 0.19).}. Hence, we refrain from attributing the lower results solely to the quality of the translated versions. It is plausible that the monolingual models themselves may exert an influence.

\section{Case Study: Presenting and Evaluating a QAPG Model} \label{sec:case_study}

In an educational context, for effective assessment, AI systems must provide not only questions but also their corresponding answer pairs. Therefore, we introduce a Question-Answer Pair Generation (QAPG) model. In this case study, our goal is to assess the extent to which QA pairs, generated using a modest-scale QAPG model and trained on translated data, can approximate real exam QA pairs and those generated using a large language model (i.e., GPT-4).
The case study focuses on European Portuguese (pt-PT). While not empirically demonstrated, we anticipate that the insights gained from the model evaluation will likely have broader applicability to other languages.

\subsection{Introducing QAPG}
\begin{figure}[htp]
    \centering
    \includegraphics[scale=0.5]{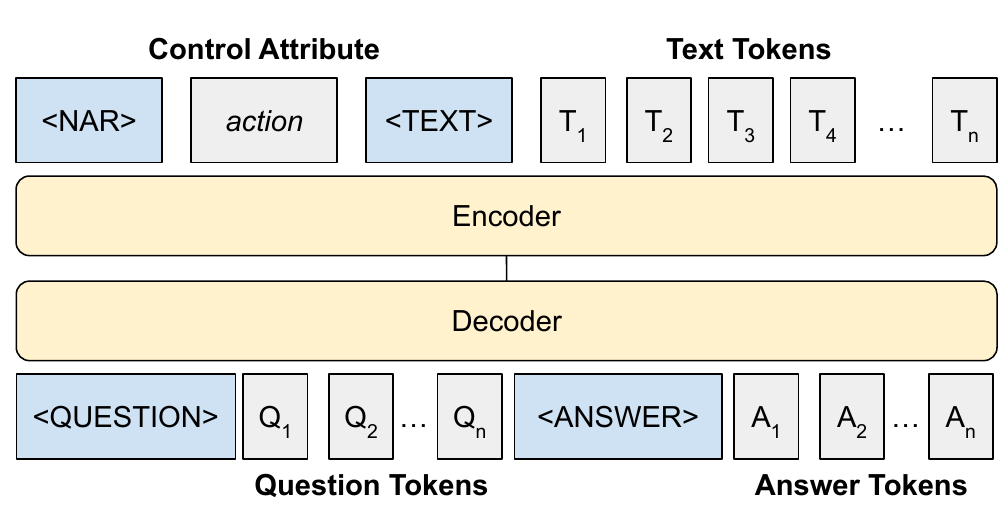}
    \caption{Setup for Question-Answer Pair Generation (QAPG), based on \cite{leite_2024_csedu}.} 
    \label{fig:qapg_overview}
\end{figure}

We frame QAPG as a controllable QA pair generation task, where narrative labels (recall Section~\ref{sec:about_dataset}) are used as control attributes. Figure~\ref{fig:qapg_overview} illustrates the QAPG model setup for generating a QA pair targeting the \textit{action} narrative element. 
The control attribute \textsc{$\langle$nar$\rangle$} is added at the start of the input, preceding the section text. The decoder outputs labels that differentiate the \textsc{$\langle$question$\rangle$} and \textsc{$\langle$answer$\rangle$} components of the output. This is based on and supported by recent studies \cite{ghanem_2022_cqg_acl,leite_2023_aied,leite_2024_csedu}, where the idea is to guide the model to generate a QA pair of the intended type.
Indeed, the controllable process is reported to hold significant importance in the educational field \cite{kurdi_2020_education}, as it enhances the creation of customized questions tailored to a student's specific needs. In our case study, we explore using different control attributes to generate diverse and multiple QA pairs from the same text.
Implementation details are similar to those explained in Section~\ref{sec:implementation_details}.

\subsection{Evaluation Protocol}
To evaluate the quality of the generated QA pairs, we engaged a group of 15 volunteers who are native speakers of Portuguese with a higher education degree.
Each individual was instructed to read a real-exam narrative text and rate a set of 15 mixed QA pairs about that text. The set included 5 QA pairs directly extracted from a real exam, 5 generated by the QAPG model\footnote{The 5 questions were randomly selected from a pool of 7, corresponding to the 7 narrative elements that can be controlled.}, and 5 generated by the GPT-4\footnote{We prompt GPT-4 Turbo to generate QA pairs from the narrative text in the target language, tailored for children aged 7 to 10, aligned with the FairytaleQA's audience.} model. We conducted 4 sets of inquiries, totaling 60 QA pairs evaluated. The texts were sourced from real exams provided by the Portuguese Ministry of Education\footnote{\url{https://iave.pt/provas-e-exames/provas-e-exames/provas-de-afericao-eb/}} for elementary school students, aligning with FairytaleQA's target audience.
For each question, participants were instructed to assess it based on the following metrics:
\begin{itemize}
    \item \textbf{Well-formedness} (question only): Is this question well-formed? [Yes: well-formed with no errors; No: ill-formed with orthographic, grammatical or semantic errors].
    \item \textbf{Relevance with Text} (question only): Is the question grounded in the text? [Yes; No].
    \item \textbf{Children Suitability} (question only): Do you consider the question suitable for a child between 7 and 10 years old? [5-point Likert scale].
    \item \textbf{Answerability} (question only): Is there an answer to the question in the text? [Yes; No].
    \item \textbf{Answer Alignment} (QA pair): Examine this potential answer: \textsc{generated answer} (\textit{the evaluators have seen the generated answer}). Is this answer valid for the question? [Yes; No].
    \item \textbf{Observations} (optional): Please justify/comment on your rating.
\end{itemize}
We ensured that at least 3 participants evaluated each QA pair.

\subsection{Results} \label{sec:results}

Table~\ref{tab:eval_60} displays the results, with each cell represented as XX/YY, where XX is the count of QA pairs with a majority positive voting result, and YY with a majority negative voting result\footnote{In 4\% of the cases, votes were tied, leading us to seek an additional volunteer to break the tie.}. An exception is the suitability metric, which is an overall average. The subsequent analysis focuses on identifying questions where, for certain metrics, the resulting vote was negative. We also emphasize cases where a positive vote result did not exhibit unanimity (100\% agreement).

\begin{table}[!ht]
\centering
\caption{Human evaluation results, determined by majority vote, except for the suitability metric, which represents an overall average.}
\label{tab:eval_60}
\begin{tabular}{c|ccc|}
\cline{2-4}
                                           & \multicolumn{3}{c|}{\textbf{QA Pairs Provenance}}                                             \\ \hline
\multicolumn{1}{|c|}{\textbf{Metric}}      & \multicolumn{1}{c|}{\textbf{Real-Exam}} & \multicolumn{1}{c|}{\textbf{GPT-4}} & \textbf{QAPG} \\ \hline
\multicolumn{1}{|c|}{Well-formedness}      & \multicolumn{1}{c|}{20/0}               & \multicolumn{1}{c|}{19/1}           & 20/0          \\ \hline
\multicolumn{1}{|c|}{Relevance with Text}  & \multicolumn{1}{c|}{20/0}               & \multicolumn{1}{c|}{20/0}           & 19/1          \\ \hline
\multicolumn{1}{|c|}{Answerability}     & \multicolumn{1}{c|}{20/0}               & \multicolumn{1}{c|}{20/0}           & 14/6          \\ \hline
\multicolumn{1}{|c|}{Answer Alignment}  & \multicolumn{1}{c|}{18/2}               & \multicolumn{1}{c|}{20/0}           & 8/12          \\ \hline
\multicolumn{1}{|c|}{Children Suitability} & \multicolumn{1}{c|}{4,77}               & \multicolumn{1}{c|}{4,83}           & 4,68          \\ \hline
\end{tabular}
\end{table}

For \textbf{well-formedness}, the questions generated by QAPG were all rated positively, showcasing the model's ability to produce questions without orthographic or grammar errors.
Two positive cases did not exhibit unanimity. Evaluators who disagreed offered valuable feedback, suggesting changes in the main verb tense for enhanced clarity. For example\footnote{The following examples have been translated from Portuguese to English.}, ``What \textit{will} the boy ask...?'' can be improved to ``What \textit{did} the boy ask...?''.
For GPT-4, one question was considered ill-formed, and another lacked unanimity. Both were identified as written in a different variant: pt-BR (instead of pt-PT).
This limitation in GPT-4 motivates further exploration, as our prompt specifically included the instruction to generate QA pairs in the intended language variant.

For \textbf{relevance with text}, despite the overall positive score, 1 question generated by QAPG received a negative vote. This question inquired about the feelings mistakenly attributed to a wrong character in the story. There were 3 questions lacking unanimity. In the first two cases, evaluators pointed out semantic inconsistencies in the context of the story, as seen in ``Who paid for the hard-boiled egg?'', which cannot be resolved based on the story. Additionally, one question was raised regarding its lack of coherence with the chronological order of events described in the text.

For \textbf{answerability}, 6 questions generated by QAPG were voted negatively, indicating that, according to the evaluators, the generated questions have no answer in the text. Upon analysis, we identified that 1 question was flagged since the required answer pertains to an event that is unclear in the story. Additionally, 2 questions were noted to demand answers that are not explicitly provided in the text. Finally, 3 questions were reported to be fully unanswerable in the text, such as, ``Who was the bear?'', a character whose description is missing in the text.

For \textbf{answer alignment}, which assesses whether the generated answer is valid for the question, QAPG shows more pronounced negative results. The first 6 negative cases are the ones already spotted for lack of answerability. In the remaining 6 cases, the generated answer was reported as inaccurate (2 cases), incomplete (1 case), or wrong/nonsensical (3 cases).
Surprisingly, 2 real-exam answers were evaluated as incorrect. Upon analysis, we detected that these answers concern a temporal location, i.e., ``The basket appeared a few days later.'' and a spatial location, ``The bear lives in the middle of nowhere.'' which are not very specific but not necessarily wrong.

Overall, the evaluators considered the questions to be \textbf{suitable} ($\geq$ 4.6) for a child aged between 7 and 10. 

\section{Discussion and Future Work}
We revisit our research question: \textit{Can a modest-scale QAPG model, trained on translated data, generate QA pairs that are qualitatively similar to those used in real exams in a less-resourced language?}

We found that the QAPG model, trained on translated data, is capable of generating well-formulated QA pairs. Although some suggestions for improvement were noted, no orthographic or grammatical errors were spotted. This is encouraging, particularly given that the dataset on which the model was trained was obtained through machine translation. Consequently, the QAPG's ability to generate well-formed questions is on par with those from real exams and GPT-4.
Then, we found that the QAPG model can generate questions that are relevant to the text. However, 1 question was reported as irrelevant, and 3 questions showed a high potential for issues related to coherence and semantics. Consequently, QAPG's ability to generate relevant questions is estimated to be slightly lower than that of questions from real exams and GPT-4.
We also discovered that 14/20 questions generated by the QAPG model were reported as answerable. Consequently, the QAPG's ability to generate answerable questions is estimated to be also slightly lower than that of questions from real exams and GPT-4.
Ultimately, we identified the primary challenge of QAPG to be the generation of aligned QA pairs. Only 8/20 of the generated answers can be considered valid to the corresponding questions. This marks a significant difference when compared to the alignment observed in real exams and GPT-4. Consequently, the QAPG's ability to generate aligned QA pairs is estimated to be significantly lower than that of questions from real exams and GPT-4. 
Here, we summarize the main issues observed when applying our QAPG model:
\begin{itemize}
\item Improper verb tense in the generated question;
\item Incoherent or semantically inconsistent generated question within the context of the story;
\item Generated question lacks an answer within the text;
\item Generated question is not aligned with the generated answer.
\end{itemize}
Overall, the results represent a preliminary yet encouraging demonstration of the potential of QA/QG systems trained on translated educational data for less-resourced languages. Despite the progress, numerous challenges persist. Considering the above points, we identify the following areas for future work as promising:
\begin{itemize}
    \item \textbf{Exploring alternative modest-scale models}: As we observed in Section~\ref{sec:benchmarks}, the performance of the monolingual QA/QG model likely influences the results. Considering recent efforts to enhance accessibility to large language models through techniques like quantization, we consider exploring alternative recent models to assess their impact on results.
    \item \textbf{Double-checking answer existence:} After generating the QA pairs, some questions were identified as having no answer in the text. One approach to confirming answer existence is by applying an additional and robust question-answering model. Comparing the answer from this model with the initially generated answer in the QA pair may provide a strong indication of answer presence.
    \item \textbf{Employing QA pair alignment verification models}: The primary challenge identified in our results lies in generating QA pairs that exhibit alignment. We suggest exploring classification models capable of classifying QA pairs as aligned or not. This approach can facilitate the filtering of unwanted QA pairs, a procedure that we find lacking in the existing QA/QG literature. Double-checking answer existence can also be useful here.
    \item \textbf{Comparing between synthetic and translated data}: GPT-4 exhibited notable superiority over the modest-scale model trained on translated data. To explore potential data-related problems, we consider conducting an experiment: employ a large language model to generate QA pairs (via zero or few-shot prompting) in the target language and then train modest-scale models on these QA pairs. This allows for a comparison between models trained on synthetic data and those trained on translated QA pairs.
\end{itemize}

\section{Conclusion}
In this work, we have advanced the accessibility of machine-translated versions of the FairytaleQA dataset, which is useful for assessing reading comprehension skills. Additionally, we developed and analyzed baseline QA/QG models. In our case study, focusing on a model trained to generate QA pairs from translated data, we successfully demonstrated the generation of well-formulated questions. However, the primary challenge lies in ensuring effective alignment within QA pairs. Although our work is limited to analyses in a single language, we hope our insights will prove valuable for researchers aiming to replicate and enhance our methods in alternative languages. For immediate future work, we intend to develop a QA pair alignment verification model to filter unwanted QA pairs.

\section*{Acknowledgments}
The authors thank the 15 participants who voluntarily participated in the evaluation process. We would like to specifically acknowledge those who agreed to share their names: \^Angela Esteves, Bruno Miguel Pinto, David Reis, Diana Pinto, Maria de Abreu, Mariana Coelho, Pedro Costa, Rui Leixo and V\'itor Magalh\~aes. This work was financially supported by Base Funding - UIDB/00027/2020 and Programatic Funding - UIDP/00027/2020 of the Artificial Intelligence and Computer Science Laboratory – LIACC - funded by national funds through the FCT/MCTES (PIDDAC). Bernardo Leite is supported by a PhD studentship (with reference 2021.05432.BD), funded by Funda\c{c}\~{a}o para a Ci\^{e}ncia e a Tecnologia (FCT).

%
%
%
\bibliographystyle{splncs04}
\bibliography{mybibliography}

\end{document}